# Syntactic Variation of Support Verb Constructions

Éric Laporte[*], Elisabete M. Ranchhod[**] and Anastasia Yannacopoulou[*]
[*]IGM, University of Marne-la-Vallée / [**]ONSET, University of Lisbon

## Introduction

The need for a robust handling of multiword expressions (MWEs) in natural language processing (NLP) is now generally acknowledged. However, in spite of the growing awareness of the problems they pose to language technology, current techniques for processing MWEs are still less effective than those for simple words. Two of the reasons for this are the variety of the linguistic forms classified as MWEs, and the lack of linguistic knowledge with such a level of formalization that it would be exploitable in computer applications.

MWEs include a large range of different linguistic objects (G. Nunberg et al., 1994; N. Calzolari et al., 2002; A. Copestake et al., 2002; I. Sag et al., 2002), such as: (i) lexical compounds (nouns: *balance of trade*, *bull's-eye, magnetic field*; adjectives: *blow by blow, high-flying*, *well-known*; adverbs: *above all*, *in crude terms, time and again*; prepositions and conjunctions: *as far as*, *in spite of*, *in order to*); (ii) phrasal verbs (*carry out*, *give up*); (iii) fixed and semi-fixed sentences (*burn the candle at both ends*, *take the bull by the horns*); (iv) support verb constructions (*give a lecture, make a speech*).

Differently from other types of MWEs, support verb constructions (SVCs), such as *Bob est dans l'embarras* (Bob is in trouble), *Bob a donné son avis sur le sujet* (Bob gave his opinion on the subject), exhibit a high degree of syntactic variation, which is important to determine and formalize. In this article, we relate experimentations and present statistical results about a type of variations of SVCs.

In the next section, we outline definitions of SVCs and predicative nouns and specify our objectives. Section 2 describes our methodology and statistical results. The conclusion draws several consequences of these results on natural language processing.

## 1. Support verbs and predicative nouns

SVCs are a major category of MWEs and should play an important role in real-world applications: machine translation, information retrieval and extraction, question answering, summarisation… Most SVCs comprise predicative nouns (PNs), an important class of nouns. The density of PNs in texts is high: about 10% of the tokens in our journalistic corpus were recognized as PNs (see section 2). A lot of scientific and technical information in texts is conveyed by PNs.

The syntactic properties of sentences with support verbs and predicative nouns have been described, from a linguistic point of view, for a number of languages, in particular for French (J. Giry-Schneider, 1978, 1987, 2005; M. Gross, 1984; L. Danlos, 1992), Italian (A. Elia et al., 1985; A. De Angelis, 1989; S. Vietri, 1996), Portuguese (E. Ranchhod, 1989, 1990), and Korean (C.-S. Hong, 1991; K. Shin, 1994; S. Han, 2000). In the French examples:

*Bob a (prêté + accordé) une grande attention à ce détail*[1]
(Bob (paid + drew + gave) special attention to this detail)

*Jo a (fait + commis) un vol*
(Jo has (done + committed) a robbery)

the verbs *prêter*, *accorder*, *faire* and *commettre* are analyzed as support verbs: their main function is to provide inflectional and aspectual information[2], whereas the PNs *attention* and *vol* are the core element of the sentence. Such a descriptive hypothesis clarifies the invariance of meaning observed in sentences where the PNs combine with different support verbs (examples above). It also explains why noun phrases headed by PNs, such as:

[(*L'avis* + *La réaction*) *de Bob sur ce point*]$_{NP}$ *était surprenant(e)*
([Bob's (opinion + reaction) on this topic]$_{NP}$ was surprising)

preserve the lexical meaning of the corresponding full sentences with support verbs:

[*Bob a (donné son avis + eu une réaction) sur ce point*]$_S$
([Bob (gave his opinion + had a reaction) on this topic]$_S$).

This close similarity between two different forms: a sentence (with verb) and a noun phrase (without verb) distinguish clearly SVCs from other constructions, and in particular from other MWEs.
Predicative nouns also preserve the argument structure of SVCs. For instance, in:

*Nous avons acquis une bonne connaissance du domaine*
(We have acquired a good knowledge of the domain)

the complement *du domaine* (of the field) is a syntactic argument of the SVC *avons acquis une connaissance* (have acquired knowledge). The prepositions and semantic role of the syntactic argument(s) depend on the lexical content of the SVC. When a PN is employed without a support verb, it often keeps that complement:

*Notre bonne connaissance du domaine*
(Our good knowledge of the domain)

The subject of the SVC can also co-occur in such a structure, the pronoun *notre* (our) in this example. Thus, sentence-like predicate/argument structures are found inside noun phrases headed by a PN, even if there is no verb.
Grammars currently used for syntactic parsing do not model the specific syntax of SVCs. Improving the coverage of such grammars with respect to SVCs requires quantitative data about the occurrences of SVCs in real texts. However, little such data are presently available. Recent, corpus-based studies provide evidence of variations of the verb in MWEs (K. Spranger 2004; B. Villada Moirón, 2005) but do not take into account the variants where the verb is absent, and they include in their scope both verbal idioms and proper SVCs. We extracted from a general-purpose corpus of texts comprehensive statistical data about occurrences of SVCs and their formal variations.

## 2. Methodology and results

2.1. Resources

Neither SVCs nor PNs are marked in currently available unambiguously tagged corpora, though ongoing work on these constructions uses the French treebank (A. Abeillé, N. Barrier, 2004). Therefore, we experimented on a raw corpus and lexical resources. The 819,000-word corpus is made of French texts from the *Le Monde* newspaper. The choice of French is motivated by the availability of large-coverage lexical resources on SVCs for this language. The size of the corpus in raw ASCII form, without tagging, is 5 Mb.

The recognition of PNs required lexical features which we found in the Lexicon-Grammar of French, a syntactic-semantic lexicon with 43,200 entries (M. Gross, 1994). We selected the entries presently available electronically and describing PNs (J. Giry-Schneider, 1978, 1987, 2005; G. Gross, 1989; A. Meunier, 1981). Since this lexicon gives syntactic information, it takes into account not only the PNs that it describes, but also the corresponding SVCs. We excluded the SVCs in which the PN is a multiword lexical unit (MWU), e.g. *donner une poignée de main* (give a handshake), where the PN *poignée de main* (handshake) is a MWU in French[3]. We obtained a sub-lexicon of 8,372 entries.

The recognition of SVCs in raw text involved the use of a morpho-syntactic lexicon. We chose Delaf, a morpho-syntactic dictionary with 950,000 entries (B. Courtois, 1990; É. Laporte, 2005). The information on SVCs from the syntactic-semantic lexicon was inserted into the Delaf. This process involved transducer-based generation of inflected forms (M. Silberztein, 2000), since the syntactic-semantic information was attached to lemmas, whereas the morpho-syntactic information was attached to inflected forms. We used the inflection tool and transducers of Unitex[4], an open-source corpus processing system (S. Paumier, 2002).

The combinations of support verbs with PNs were recognized through finite-state parsing (E. Roche, Y. Schabès, 1997). Since no available grammar of French takes into account SVCs with sufficient coverage, we manually designed two grammars, one for SVCs and the other for PNs in general. Both grammars take the form of recursive transition networks (RTNs), but they do not make use of the possibility of recursion, and therefore they are purely finite-state. The two grammars total 269 graphs. They were created and applied with the aid of Unitex.

The PN grammar recognizes PNs with some left context, in order to resolve lexical ambiguity and improve precision. For example, *débat* (debate) is ambiguous with a form of the verb *débattre* (to debate). The PN grammar resolves the ambiguity by recognizing it only when in association with a determiner in grammatical agreement with it, as in *ce débat* (this debate), which is unambiguous. The context recognised by the PN grammar consists of determiners, the prepositions *de* (of) and *sans* (without), which can be employed without determiner, and some punctuation marks, such as opening parenthesis. The description of determiners is inspired from M. Gross (2001) and M. Silberztein (2003), but has been entirely reorganized.

The SVC grammar recognises constructions with a PN and the associated support verb, either placed before, as in *donne l'explication* (give an explanation), or placed after, as in *un entretien (a été) accordé* (an interview (was) given). It recognises 70 support verbs (see Appendix).

## 2.2. Classification of occurrences

The PN grammar identified 95,430 occurrences of PNs. With the SVC grammar, we classified them into two groups, depending on the presence vs. absence of the support verb. Only 3,349 of the occurrences recognised by the PN grammar were also recognised by the SVC grammar.

Thus, only about 4% of occurrences of PNs are accompanied by their associated support verb. This tends to establish that, in most cases, PNs are not associated to an explicit occurrence of a support verb. Recall that there is no clear-cut difference of meaning between occurrences of the same PN with vs. without a support verb (cf. section 1). A full SVC and the same PN occurring without a support verb are clearly variants of a single syntactic-semantic object.

## 2.3. Assessment of biases

Several biases affect our experimentations and can distort the correctness of our statistics.

The first bias is the incompleteness of the syntactic-semantic lexicon of PNs described in 2.1. This incompleteness stems from three facts: we excluded multiword PNs; some parts of the Lexicon-Grammar of French, e.g. the lexicon of disease nouns (J. Labelle, 1986), are not available electronically; some sub-categories of PNs, e.g. those that are in subject position in the SVC (e.g. *un phénomène a eu lieu* (a phenomenon took place)), have not been formalised, to our knowledge, by the Lexicon-Grammar authors. The incompleteness of the lexicon has an impact on recall in our recognition experimentations.

The second bias is the incompleteness of the grammars. Some syntactic constructions are lacking. For instance, we did not include the constructions in which the support verb occurs in a relative clause attached to the PN: *les répercussions qu'aura l'événement* (the repercussions that the event will have). Some variants of support verbs are not taken into account, e.g. *établir* (establish) as a variant of *faire* (make) associated to the PN *grille* (table): *une grille préalablement établie* (a previously established table). The absence of constructions from the grammars affects recall. Symmetrically, not all syntactic constraints are formalised in our grammars; in particular, they are lexicalised at the level of sub-categories of PNs, not at the level of individual lexical entries. Consequently, not all syntactic properties of each PN are taken into account. For instance, our grammar recognises *présentent les principales notions* (present the main notions) as an instance of the SVC *avoir une notion de* (have a notion of), though with this PN, *présenter* (to present) is not an acceptable variant of *avoir* (have)[5]. The absence of formalisation of some constraints affects precision.

The third bias stems from lexical ambiguity. Unitex performs lexical tagging with the aid of morpho-syntactic lexicons. Recall was 99.3%, but several solutions per word are retained in case of ambiguity, which causes mismatches. For instance, *les nouvelles données* (the new data) is recognised as an instance of the SVC *donner des nouvelles* (give news), because both *nouvelles* and *données* are lexically ambiguous. In the correct interpretation, *nouvelles* is an adjective (new) and *données* a noun (data). In the SVC interpretation, *nouvelles* is a plural noun (news), and *données* is a support verb (given), as in *Les nouvelles données par Marie étaient surprenantes* (The news given by Mary were surprising). This bias has an impact on precision.

The last bias is caused by errors in the corpus, e.g. *pilliers* for *piliers* (pillars). This bias is generally considered negligible by users of corpora of texts from this newspaper, since it applies a highly effective correction process to its texts.

In order to assess the distortions caused by all four biases, we carried out additional experimentations on two sub-corpora of different sizes constituted by paragraphs extracted arbitrarily from the corpus of 2.1. Sub-corpus C1 contains 6,850 words. Sub-corpus C2 contains 205,000 words. Two trained experts, E1 and E2, annotated manually the occurrences of PNs and SVCs. We compared the output of their work with the concordances produced by Unitex and we computed the recall and the precision.

E1 and E2 adopted common guidelines in their judgments on the occurrence of PNs and SVCs:

(i) Exclude from SVCs the phrases in which a semantically full verb is in a syntactically and semantically compositional combination with a noun, e.g. *(cherche à) acquérir les actions* ((tries to) buy the stocks).

(ii) Exclude from SVCs the verbal idioms containing a verb which is present in all syntactic variants of the idiom, e.g. *pour faire face à (une telle crise)* (to cope with (such a crisis)). In all the syntactic variants of the phrase *faire face* with the meaning (cope with), both words are simultaneously present. Without *faire*, the noun *face* has different meanings (side, face, dignity).

(iii) Exclude SVCs with a preposition frozen with the noun phrase, e.g. *qui vont à contre-courant* (that go counter-current). In this phrase, the metaphorical meaning is observed only with the preposition *à* [6].

(iv) Exclude from PNs nouns denoting professions, e.g. *avocat* (lawyer), and parts of objects, e.g. *aile* (wing). Such expressions as *avoir un avocat* (have a lawyer) and *avoir une aile* (have a wing) are sometimes considered as predicative forms.

(v) Include multiword PNs, e.g. *poignée de main* (handshake).

(vi) Include PNs which can occur as the subject of their support verb, e.g. *bouleversement* (upheaval) which can occur in *Un bouleversement a eu lieu* (An upheaval took place).

The recall scores were evaluated on sub-corpus C1 through the following procedure. E1 found 646 PN occurrences. We compared the output of this work with the concordance sorted according to the text order, and counted the concordance lines that matched with occurrences marked by E1. We computed the recall on these values. E2 proceeded in the same way in parallel. We computed the average of the two recall scores obtained. We proceeded in the same way for SVCs. Table 1 shows the results[7].

|            | PNs  | SVCs |
|------------|------|------|
| E1         | 646  | 48   |
| E1 & Unitex| 564  | 28   |
| Recall     | 87%  | 58%  |
| E2         | 820  | 85   |
| E2 & Unitex| 561  | 17   |
| Recall     | 68%  | 20%  |
| Average    | 78%  | 38%  |

Table 1. Recall scores

The precision scores were evaluated as follows. For the precision of the recognition of PNs we used sub-corpus C1. We computed the precision scores from the count in Table 1 (E1 & Unitex) and the total number of concordance lines. We did the same for E2 and we computed the average of the two precision scores. For the precision of the recognition of SVCs, we used sub-corpus C2 in order to obtain more reliable values. E1 annotated manually the occurrences of actual SVCs in the Unitex concordance of this sub-corpus. We counted these occurrences and computed the precision from this count. E2 proceeded in the same way in parallel. We computed the average of the two precision scores. Table 2 shows the results.

|              | PNs  | SVCs |
|--------------|------|------|
| Unitex       | 831  | 895  |
| E1 & Unitex  | 564  | 751  |
| Precision    | 68%  | 84%  |
| E2 & Unitex  | 561  | 576  |
| Precision    | 68%  | 64%  |
| Average      | 68%  | 74%  |

Table 2: Precision scores

The precision of SVC recognition is higher than that of PN recognition, because lexical ambiguity has less influence on the recognition of longer sequences.

We extrapolated these values of recall and precision to the whole corpus in order to correct the results of 2.2. We applied the following correction formula to the number of occurrences of PNs obtained in 2.2:

$$n' = np/r$$

where $n$ is the experimental value, $n'$ the corrected value, $p$ the precision and $r$ the recall. We did the same for the number of occurrences of SVCs. The experimental results and the results corrected on the basis of the extrapolation are displayed in Table 3.

|              | PN    | SVC  | Proportion |
|--------------|-------|------|------------|
| experimental | 95430 | 3349 | 4%         |
| corrected    | 83195 | 6522 | 8%         |

Table 3: Corrections to the results of 2.2.

In the corrected results, the occurrences of PNs inside a SVC are slightly more numerous but remain clearly a minority.

Finally, in order to validate our assessment of recall and precision, we compared the lists of occurrences respectively marked as PNs and as SVCs by the two annotators. As it turned out, the differences between E1 and E2 come from differences of appreciation in the application of the 6 guidelines above. In particular, guidelines (i) and (ii) involve criteria which are sometimes difficult to check, especially with sub-categories of PNs which have not been submitted to systematic linguistic studies yet. For example, relational nouns such as *représentant* (representative) easily combine with *avoir* (have), but the annotators often disagreed on whether the construction corresponds to a SVC or to a compositional verb phrase (cf. (i)). The boundary between SVCs and verbal idioms is also sometimes difficult to determine, e.g. in *Le groupe a <u>fait du chemin</u> depuis sa première apparition* (This group <u>made way</u> since their first show).

In the last analysis, these differences of judgment between E1 and E2 reflect uncertainty about the precise limits of PNs and SVCs. The fact that the agreement between E1 and E2 is better as regards precision than as regards recall suggests that this uncertainty is larger for sub-categories of PNs not described in literature.

Therefore, our evaluation of recall and precision are clearly approximations, but we consider unlikely that experimentation with other experts or on other texts would yield results of a different order of magnitude.

2.4. Results by sub-categories

The 8,372 PNs in our syntactic-semantic lexicon have quite a variety of behaviours. We wanted to know whether the results above depend on sub-categories. We used the sub-categorization provided by this lexicon. The first sub-category (CV) includes the PNs that enter in a converse construction (G. Gross, 1989), such as *Marie fait un baiser à Max* (Mary gives Max a kiss) and *Max reçoit un baiser de Marie* (Max gets a kiss from Mary). The second sub-category (NCF) includes the PNs that accept the support verb *faire* (do/make) but do not enter in a converse construction, for instance *Marie fait une promenade* (Mary takes a walk). Finally, the PNs that admit the support verb *avoir* 'have' and do not enter in a converse construction, such as *Marie a une idée* (Mary has an idea), constitute sub-category NCA. In order to see how the proportion found in 2.2 depends on these three sub-categories, we made three versions of our PN grammar dedicated to each of these sub-categories. We did the same for the SVC grammar. We applied these grammars to the complete corpus. We obtained data about the relative numerical importance of the sub-categories. In Table 4, we show in the 'PN%' line the proportion of PN occurrences which are recognized by the respective grammars.

|           | NCA   | NCF   | CV    | all   |
|-----------|-------|-------|-------|-------|
| PNs       | 56457 | 42420 | 30231 | 95430 |
| PN %      | 59%   | 44%   | 32%   | 100%  |
| SVCs      | 1600  | 868   | 1334  | 3349  |
| SVC %     | 48%   | 26%   | 40%   | 100%  |
| SVC/PN    | 3%    | 2%    | 4%    | 4%    |
| corrected | 6%    | 4%    | 10%   | 8%    |

Table 4: Results by sub-categories

The sum of percentages exceeds 100%, because some occurrences are recognized by more than one grammar[8]. The same information is provided for SVCs. The sub-categories do not have the same statistical significance in text. The last two lines show the proportion of PNs that are recognised by the SVC grammars, as in Table 3. The corrected values are computed with the method presented in 2.3.

The percentages depend on the sub-categories but remain in the same order of magnitude.

**Conclusion**

We provided empirical evidence that most PNs (approximately 92%) are not associated with an explicit occurrence of a support verb[9]. This important underlying property of SVCs[10] is in general not shared by other MWEs. In general, a MWE is composed of at least two elements which are simultaneously present, even if they may undergo variations. For instance, *dorer la pilule* (to sugar the pill) admits a passive form: *la pilule était dorée* (the pill was sugared), but the specific meaning of the idiom appears only in the presence of both a form of *dorer* and a form of *pilule*.

This difference has a consequence on computational treatment: computational techniques for extracting or recognising other MWEs will probably not easily transfer to SVCs. Thus, it appears important to make the distinction between SVCs and verbal idioms.

Our work shows how linguistic notions elaborated through manual description can be useful for an NLP-oriented typology of MWEs. In a similar way, the availability of large-scale, manually constructed lexical resources should offer valuable opportunities to annotate corpora on the basis of linguistic analyses and to train statistical models on them.

**References**

Abeillé, Anne; Nicolas Barrier. 2004. Enriching a French treebank. *Proceedings of LREC*, Lisbon.

Calzolari, Nicoletta; Charles Fillmore; Ralph Grishman; Nancy Ide; Alessandro Lenci; Catherine MacLeod; Antonio Zampolli. 2002. Towards Best Practice for Multi-Word Expressions in Computational Lexicons. In *Proceedings of LREC*, Las Palmas, 1934-1940.

Copestake, Ann; Fabre Lambeau; Aline Villavicencio; Francis Bond; Timothy Baldwin; Ivan Sag; Dan Flickinger. 2002. Multiword expressions: linguistic precision and reusability. *Proceedings of LREC*, Las Palmas, 1941-1947.

Courtois, Blandine. 1990. Un système de dictionnaires électroniques pour les mots simples du français, *Langue Française* 87: 11-22.

Danlos, Laurence. 1992. Support Verb Constructions: linguistic properties, representation, translation. *Journal of French Language Studies* 2(1): 1-32.

De Angelis, Angela. 1989. Nominalizations with the Italian support verb *avere*. *Lingvisticae Investigationes* 13(2):223-238.

Elia, Annibale; Emilio D'Agostino; Maurizio Martinelli. 1985. Tre componenti della sintassi italiana: frasi semplici, frasi a verbo supporto e frasi idiomatiche. *Proceedings of the 17th International Conference of SLI*, Roma: Bulzoni, 311-325.

Giry-Schneider, Jacqueline. 1978. *Les nominalisations en français. L'opérateur « faire » dans le lexique*, Genève: Droz.

Giry-Schneider, Jacqueline. 1987. *Les prédicats nominaux en français. Les phrases simples à verbe support*, Genève: Droz.

Giry-Schneider, Jacqueline. 2005. Les noms épistémiques et leurs verbes supports, *Lingvisticæ Investigationes* 27(2):219-238.

Gross, Gaston. 1989. *Les constructions converses du français*, Genève: Droz.

Gross, Maurice. 1984. Lexicon-Grammar and the Syntactic Analysis of French. *Proceedings of Coling*, Stanford, 275-282.

Gross, Maurice. 1994. Constructing Lexicon-grammars. *Computational Approaches to the Lexicon*, Atkins and Zampolli (eds.), Oxford University Press, 213-263.

Gross, Maurice. 2001. Grammaires locales de déterminants nominaux. *Détermination et Formalisation*, Blanco, Buvet, and Gavriilidou (eds.), *Lingvisticae Investigationes Supplementa* series, 23, 177-193.

Han, Sun-hae. 2000. *Les prédicats nominaux en coréen : Constructions à verbe support* hata, PhD Thesis, University of Marne-la-Vallée.

Hong, Chai-Song. 1991. Objet interne, verbe support et dictionnaire. *Proceedings of Computational Lexicography (Balatonfüred, Hungary, 1990)*, Budapest: Hungarian Academy of Sciences, 93-102.

Labelle, Jacques. 1986. Grammaire des noms de maladie. *Langue Française* 69:108-128.

Laporte, Éric. 2005. Symbolic Natural Language Processing. *Applied Combinatorics on Words*, Lothaire, Cambridge University Press, 164-209.

Meunier, Annie. 1981. *Nominalisations d'adjectifs par verbes supports*. PhD thesis, LADL, University of Paris 7.

Nunberg, Geoffrey, Ivan Sag, and Thomas Wasow. 1994. Idioms. *Language* 70(3):491-538.

Paumier, Sébastien. 2002. *Unitex. Manuel d'utilisation*, Research report.

Ranchhod, Elisabete. 1989. Predicative Nouns and Negation. *Lingvisticae Investigationes* 13(2):387-397.

Ranchhod, Elisabete. 1990. *Sintaxe dos predicados nominais com* estar, Lisbon: INIC.

Roche, Emmanuel; Yves Schabès (eds.). 1997. *Finite-State Language Processing*. Cambridge, Massachusetts: MIT Press.

Sag, Ivan; Timothy Baldwin; Francis Bond; Ann Copestake; Dan Flickinger. 2002. Multiword Expressions: a Pain in the Neck for NLP. *Proceedings of CICLing*, Mexico, 1-15.

Shin, Kwang-soon. 1994. *Le verbe support* hata *en coréen contemporain : morpho-syntaxe et comparaison*, PhD thesis, University of Paris 7.
Silberztein, Max. 2000. Intex: an FST toolbox. *Theoretical Computer Science*, 231(1): 33-46.
Silberztein, Max. 2003. Finite-State Description of the French Determiner System. *French Language Studies* 13:221-246.
Spranger, Kristina. 2004. Beyond subcategorization acquisition. Multi-parameter extraction from German text corpora. *Proceedings of Euralex*, I:171-177.
Vietri, Simonetta. 1996. The Syntax of the Italian verb *essere Prep. Lingvisticae Investigationes* 20(2):287-363.
Villada Moirón, Begoña. 2005. Linguistically enriched corpora for establishing variation in support verb constructions. *Proceedings of LINC*, Jeju Island, Republic of Korea.

**Abstract - Syntactic Variation of Support Verb Constructions**

We report experiments about the syntactic variations of support verb constructions, a special type of multiword expressions (MWEs) containing predicative nouns. In these expressions, the noun can occur with or without the verb, with no clear-cut semantic difference. We extracted from a large French corpus a set of examples of the two situations and derived statistical results from these data. The extraction involved large-coverage language resources and finite-state techniques. The results show that, most frequently, predicative nouns occur without a support verb. This fact has consequences on methods of extracting or recognising MWEs.

**Appendix - List of the support verbs included in the SVC grammar**

| | | | | | | | |
|---|---|---|---|---|---|---|---|
| *accorder* | *causer* | *encaisser* | *faire* | *manquer* | *porter* | *provoquer* | *retenir* |
| *acquérir* | *commettre* | *encourir* | *filer* | *multiplier* | *pratiquer* | *réaliser* | *s'adonner* |
| *administrer* | *concevoir* | *endurer* | *fixer* | *nourrir* | *prendre* | *recevoir* | *se livrer* |
| *adresser* | *connaître* | *entamer* | *flanquer* | *obtenir* | *prescrire* | *redonner* | *subir* |
| *allonger* | *conserver* | *entreprendre* | *garder* | *octroyer* | *présenter* | *refaire* | *tenir* |
| *apporter* | *déborder* | *éprouver* | *imposer* | *offrir* | *préserver* | *regorger* | *tirer* |
| *asséner* | *donner* | *essuyer* | *infliger* | *passer* | *procéder* | *reperdre* | *toucher* |
| *avoir* | *écoper* | *être* | *jeter* | *percevoir* | *procurer* | *reprendre* | |
| *caresser* | *effectuer* | *exercer* | *lancer* | *perdre* | *prononcer* | *ressentir* | |

*Authors' addresses:*

Éric Laporte [1]    Elisabete Marques Ranchhod [2]    Anastasia Yannacopoulou [1]
[1]IGM, University of Marne-la-Vallée          [2]ONSET, University of Lisbon
5, bd Descartes                                Alameda da Universidade
77454 Marne-la-Vallée CEDEX 2                  1600-214 Lisboa
France                                         Portugal

eric.laporte@univ-mlv.fr
e_ranchhod@fl.ul.pt
anastasia.yannacopoulou@univ-mlv.fr

[1] Textual elements that can commute are written in parentheses, separated by a + sign.

[2] Some support verbs regularly convey aspectual information: cf. *Le phénomène (a + prend) des proportions inquiétantes* (This phenomenon (assumes + is reaching) worrying proportions). Others convey a stylistic difference: cf. *Les deux versions (avaient + présentaient) des avantages* (Both versions (had + presented) advantages).

[3] This is because the tool we used to inflect PNs was operational only on simple words, not on MWUs.

[4] http://igm.univ-mlv.fr/~unitex/

[5] In other SVCs, such as *avaient des avantages* (had advantages), *présentaient* (presented) is an acceptable variant of the support verb *avaient* (had).

[6] This is a different situation from, for instance, the SVC *qui procédait à un banal contrôle d'identité* (who was performing a routine identity check), where *procédait à* is a variant of *faisait*: *qui faisait un banal contrôle d'identité* (who was doing a routine identity check), and *à* is not frozen with the PN.

[7] Most of the silence in the recognition of PNs is caused by the incompleteness of the syntactic-semantic lexicon. If we exclude from the PN occurrences marked by the annotators those not described in this lexicon, not even with another meaning, the recall score rises to 96% for E1, and to 94% for E2. In the case of the SVC occurrences, the corresponding values are 70% and 25%, which suggests that the silence is also caused, in that case, by the incompleteness of the grammar.

[8] For example, the PN *pêche* (fishing) accepts the support verb *faire* (do) and is subcategorized as NCF, whereas *pêche* (punch (colloquial)) is used with the support verb *avoir* (have) and belongs to CV.

[9] As a result, PNs co-occur not only with support verbs, but also with semantically full verbs, some of which are ambiguous with support verbs, e.g. *présenter* (to present). This is an obstacle to the automatic identification of SVCs with the aid of co-occurrence statistics (Spranger, 2004; Villada Moirón, 2005).

[10] Our experiment was carried out on French data only. However, as regards English and all the languages for which results of comprehensive studies on SVCs are available, the situation seems comparable. In Italian, Korean, and Portuguese, linguists describe variants of SVCs in which the PN is not associated with an occurrence of a support verb. Moreover, it is possible to identify cross-linguistically a few types of SVCs and some types of variations of SVCs which are common to these languages or to several of them. This allows us to exclude the hypothesis of a French-specific syntactic variability that would be an exception among natural languages.